\documentclass[lettersize,journal]{IEEEtran}
\usepackage{amsmath,amsfonts}
\usepackage{algorithmic}
\usepackage{algorithm}
\usepackage{array}
\usepackage[caption=false,font=normalsize,labelfont=sf,textfont=sf]{subfig}
\usepackage{textcomp}
\usepackage{stfloats}
\usepackage{url}
\usepackage{verbatim}
\usepackage{graphicx}
\usepackage{cite}
\usepackage{tabularx}
\usepackage{multirow}
\usepackage{colortbl}
\usepackage{bbding}
\usepackage{color}
\usepackage{adjustbox}
\usepackage{booktabs}
\usepackage{subcaption}
\usepackage{hyperref}
\usepackage{orcidlink} 
\hypersetup{
    colorlinks=true,
    linkcolor=blue,
    citecolor=blue, 
    urlcolor=blue  
}

\begin{document}

\title{GA\textsuperscript{2}-CLIP: Generic Attribute Anchor for Efficient Prompt Tuning in Video-Language Models}

\author{Bin Wang$^{\orcidlink{0009-0002-8144-1280}}$, Ruotong Hu, Wentong Li, Wenqian Wang, Mingliang Gao$^{\orcidlink{0000-0001-7273-7499}}$, Runmin Cong$^{\orcidlink{0000-0003-0972-4008}}$, \textit{Senior Member, IEEE},\\ Wei Zhang$^{\orcidlink{0009-0008-5809-1335}}$, \textit{Senior Member, IEEE}, Xudong Jiang$^{\orcidlink{0000-0002-9104-2315}}$, \textit{Fellow, IEEE}

\thanks{Bin Wang is with the Shandong Provincial Key Laboratory of New Power Distribution \& Utilization Technology and Equipment, School of Electrical and Electronic Engineering, Shandong University of Technology, Zibo 255000, China, also with the School of Control Science and Engineering, Shandong University, Jinan 250061, China (e-mail: dqwangbin@sdut.edu.cn).

Ruotong Hu is with the School of Computer Science and Technology, Shandong University of Technology, Zibo 255000, China (e-mail: hrt@sdut.edu.cn).

Wentong Li is with the School of Artificial Intelligence, Nanjing University of Aeronautics and Astronautics, Nanjing 210000, China (e-mail: wentong\_li@nuaa.edu.cn).

Wenqian Wang is with the Pillar of Information Systems Technology and Design, Singapore University of Technology and Design, Singapore (e-mail: wenqian\_wang@sutd.edu.sg).

Mingliang Gao is with the School of Electrical and Electronic Engineering, Shandong University of Technology, Zibo 255000, China (e-mail: mlgao@sdut.edu.cn).

Runmin Cong and Wei Zhang are with the School of Control Science and Engineering, Shandong University, Jinan 250061, China, also with the Key Laboratory of Machine Intelligence and System Control, Ministry of Education, Jinan 250061, China (e-mail: rmcong@sdu.edu.cn; davidzhang@sdu.edu.cn).

Xudong Jiang is with the School of Electrical and Electronic Engineering, Nanyang Technological University, Singapore 639798 (e-mail: exdjiang@ntu.edu.sg).

(\emph{Corresponding author: Wei Zhang.})  \\
}
}



\maketitle

\begin{abstract}
Visual and textual soft prompt tuning can effectively improve the adaptability of Vision-Language Models (VLMs) in downstream tasks. However, fine-tuning on video tasks impairs the model's generalization ability to unseen classes. Existing methods attempt to mitigate this forgetting effect by regularizing the gap between hand-crafted prompts and soft prompts, but this also weakens the learning ability of soft prompts. To address this challenge, we propose a plug-and-play coupling prompt learning framework to optimize the generalization performance of V-L models in video tasks, with the core motivation of mitigating semantic space narrowing during fine-tuning by introducing an externally supervised prompt. Specifically, for textual prompts, we introduce pre-trained prompts from other datasets as hard prompt tokens. These are concatenated with soft prompt tokens and coupled via a learnable mapping layer. This competitive prompting approach prevents the semantic space from overfitting to supervised categories. In addition, we introduce a set of well-designed irrelevant video sets and negative prompts as generic attribute anchors to maintain the generic relevance of the attributes in the pre-trained semantic space, thus preserving the generalization ability. Experiments on video tasks demonstrate that our method significantly outperforms state-of-the-art prompt tuning approaches across generalization benchmarks, particularly on base-to-new class prediction. The code is available at \href{https://github.com/BBYL9413/GA2-CLIP}{https://github.com/BBYL9413/GA2-CLIP.}

\end{abstract}
    
\section{Introduction}

\begin{figure}[t]
    \centering
    \includegraphics[width=\linewidth]{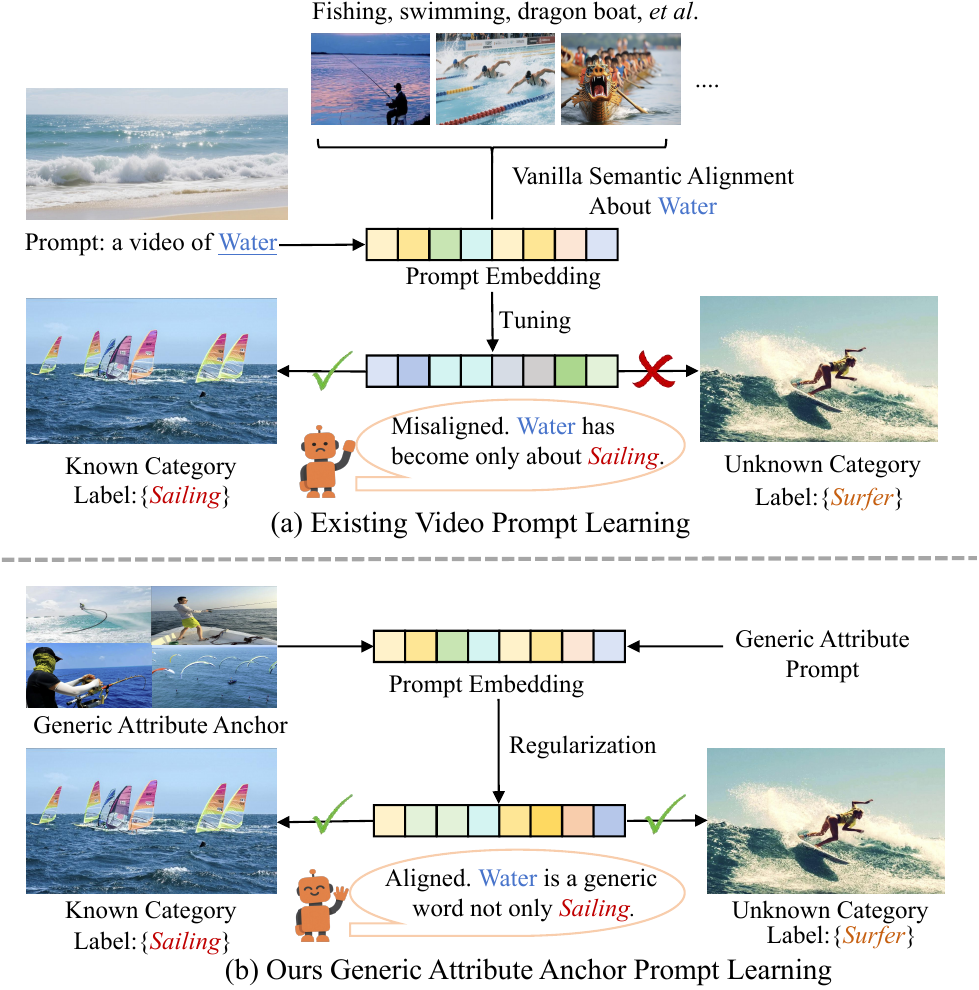}
    \caption{Comparative analysis of video-text alignment process via learnable prompts. (a) Current video prompt fine-tuning learning methods destroy the original semantic spatial information, causing the model to lose the ability to discriminate between unknown categories. (b) GA\textsuperscript{2}-CLIP mitigates video semantic bias towards known category by introducing generic attribute anchors and generic attribute prompt.}
    \label{fig1}
\end{figure}

\label{sec:intro}

Large-scale Vision-Language Models (VLMs), such as CLIP \cite{clip}, ALIGN \cite{jia2021scaling}, and Florence \cite{yuan2021florence}, demonstrate strong generalized representation learning capabilities. This capability is attributed to the large image language training data set, which uses contrast learning to achieve excellent zero-shot generalization capabilities in various image tasks. Though some image-based VLM methods such as ATPrompt~\cite{li2024atprompt}, DCP~\cite{li2025dpc}, and ArGue~\cite{tian2024argue} have recently achieved impressive results, however this model is difficult to replicate perfectly in video-related tasks, the computational cost of training a video-language model increases exponentially, and it is difficult to align all the content in a video through language. Therefore, recent researchers are trying to migrate prompt learning to video action recognition tasks, which achieves good performance with only a few learnable soft prompts\cite{jia2022visual}.


Inspired by image classification tasks \cite{khattak2023self, zhou2022conditional, xie2025textrefiner}, current text feature-based video learning methods mainly replace hand-crafted fixed text prompts by combining learnable soft prompts with hard class labeling. Although this approach can be well adapted to the downstream video task, as shown in Fig.~\ref{fig1}(a), it can easily over-align the known categories in the video thus losing the ability to generalize to unknown categories. Intuitively, this stems from semantic narrowing within the original textual semantic space during base-class video training. Taking “water” in the figure as an example: while it inherently associates with diverse motions (e.g., fishing, swimming) in the original semantic space, exposure to limited downstream video contexts involving “water” biases its semantic space toward known water-related categories (e.g., sailing). Consequently, “water” becomes a class-specific attribute. In addition, the complex background information of the video may form spurious semantic correlations \cite{tian2024argue}, thus losing the capture of critical motion information.

Building on these observations, we propose an innovative framework, GA\textsuperscript{2}-CLIP, to counter semantic narrowing in the fine-tuning process by introducing external supervision. Specifically, we construct a set of generic attribute anchors disjoint from both base and novel category, as visualized in Fig.~\ref{fig1}(b). During model learning, these anchors are aligned with generic attribute prompts through contrastive learning, and integrated into the training objective as a regularization term. Compared to predefined attribute methods \cite{li2024atprompt, wang2024vilt, zhu2023prompt}, it does not require filtering of useful attributes for complex videos; instead, the model learns these features implicitly. In addition, we introduce a set of pre-trained hard prompt tokens for supervised semantic learning of soft prompt tokens, with feature coupling via a nonlinear mapping layer. It can effectively prevent the soft prompt token from fully adapting to the label semantics of the base category, which is used as input noise to realize competitive prompt learning.


In summary, this paper focuses on how to guide the semantics of textual features to effectively correlate the key information of the video, to improve the generalization performance of the video task. GA\textsuperscript{2}-CLIP is a plug-and-play technique without additional parameters, which enables the existing models to achieve good video performance enhancement.
Our contributions are summarized as follows:
\begin{itemize}
    \item We propose a generic attribute anchor prompting approach for video language fine-tuning to counteract semantic narrowing in downstream tasks.
    \item We introduce an externally supervised hard prompt coupled with soft prompts via nonlinear mapping layers, enhancing generalization through competitive learning mechanisms.
    \item Extensive experiments demonstrate that GA\textsuperscript{2}-CLIP achieves seamless integration with existing video-text learning methods, boosting performance with negligible computational overhead.
    
\end{itemize}

\section{Related work}
\label{sec:work}

\begin{figure*}[ht]
    \centering
    \includegraphics[width=\linewidth]{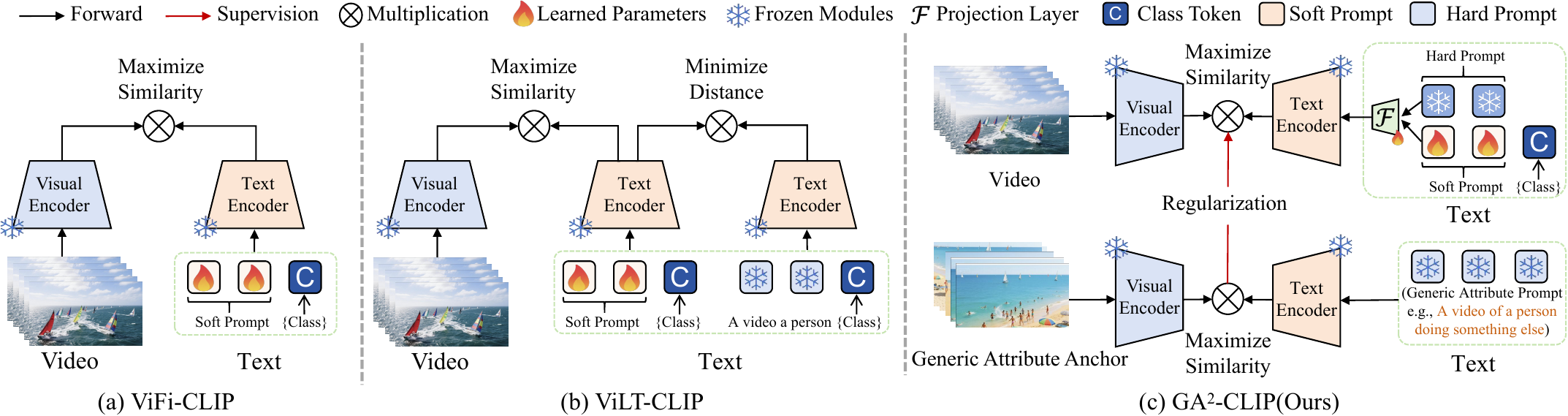}
    \caption{Comparison of existing video prompt tuning architectures. (a) ViFi-CLIP inputs multiple learnable soft tokens combined with class tokens to the text encoder. (b)ViLT-CLIP guides the soft tokens learning by introducing hand-crafted prompt templates from vanilla CLIP to avoid soft tokens learning too well in the base category and thus losing their generalization ability. (c) Our GA\textsuperscript{2}-CLIP innovatively introduces generic attribute anchors and hard prompts to guide the learning of soft tokens, and shows strong performance in both the base category and the novel category. }
    \label{fig2}    
\end{figure*}

\textbf{Vision Language Models:} The integration of language supervision with natural image in Visual-language (V-L) models \cite{jia2021scaling,zhai2022lit,liu2024llava} has emerged as a significant research focus in computer vision. Such models utilize large-scale image-text pairs for contrastive learning to construct joint feature representation spaces across modalities, with the core advantage of encoding rich semantic information and supporting zero-shot \cite{wang2024vilt} and few-shot \cite{zhang2021tip,silva2024closer} knowledge transfer. These methods enable the learning of fine-grained visual concepts through self-supervised training and show excellent generalization capabilities in downstream tasks in the image domain (e.g., semantic segmentation \cite{rao2022denseclip}, target detection \cite{du2022learning}, point cloud classification \cite{xu2024pointllm}) and video tasks (video classification, retrieval). However, V-L models in the video domain still face challenges due to issues such as insufficient data size. In this work, we present a new framework utilizing prompt fine-tuning that can cost-effectively migrate visual and language models to video tasks.


\noindent\textbf{Video Action Recognition:} Video action recognition has undergone a leap from convolutional neural networks \cite{qiu2017learning, tran2018closer,feichtenhofer2020x3d, lin2019tsm, li2020tea, wang2021tdn} to transformer \cite{arnab2021vivit, fan2021multiscale,neimark2021video,bertasius2021space,ranasinghe2022self, liu2022video} architectures, with the key to learning spatiotemporal representations from video. Recently, with the emergence of powerful vision-language pre-training models like CLIP, which show remarkable generalization across a variety of image tasks, researchers start to migrate them to video tasks. \cite{pan2022st,qing2023disentangling,yang2023aim,wang2024tds} add trainable temporal adapters to the model to achieve dynamic feature learning. \cite{zhang2020side, sung2022lst, liu2023revisiting} use ladder models to migrate pre-training weights, avoiding back-propagation of gradients thus freeing up more memory. \cite{wang2023actionclip,wu2023bidirectional, wang2024multimodal} devise multi-modal temporal modeling architectures to achieve high-performance action recognition while maintaining high robustness and transferability. Vita-CLIP \cite{wasim2023vita} and ViLT-CLIP \cite{wang2024vilt} introduce multi-modal prompt learning in both vision and text thereby balancing the generalized supervised performance for video tasks. Although prompt fine-tuning has been successfully introduced to video tasks, we note that while existing methods follow multi-modal solutions, the generalization performance still needs to be improved.


\noindent\textbf{Vision Prompt Learning:} Text prompt is a novel paradigm in the field of natural language processing that uses sentence forms as templates, and is often used in the language branches of V-L models to better help the models understand the semantics. Prompts are categorized into manual and automatic forms \cite{khattak2023maple, jia2022visual, ju2022prompting}, and automatic forms are used for the fine-tuning phase of automatic learning, often called Prompt Learning. Current language prompt learning is widely used in multi-modal models, e.g., CoOp and Co-CoOp \cite{zhou2022conditional} enhance language coherence by optimizing successive prompt vectors in language branches. PromptKD \cite{li2024promptkd} supervises the transfer of knowledge from CLIP to a lightweight model through an unsupervised approach using a teacher model.  ArGue \cite{tian2024argue} utilizes a large-scale language model to mine multiple in-class attributes and proposes a negative sample prompt to eliminate spurious connections caused by backgrounds. ATPrompt \cite{li2024atprompt} extends the learning space of soft prompts from the one-dimensional category level to the multidimensional attribute level by proposing hard attribute anchors to improve the generalization of attributes to new categories. Most of the existing research uses visual attributes to enhance model performance, which is achieved by providing additional attribute information through LLM, etc. However this is extremely difficult in video tasks, where it is difficult to correlate all the actions with the background, and therefore the model cannot be guided to learn the action features by means of predefined attributes. In this work, we rethink the significance of video attributes. We argue that while we cannot tell the model which attributes in a video are positive, we are able to guide the model on which attributes may be meaningless. We implement video learning for the model by proposing a global generic attribute anchor and a local hard-supervised prompt.
\section{Methodology}
\label{sec:method}

\begin{figure*}[ht]
    \centering
    \includegraphics[width=\linewidth]{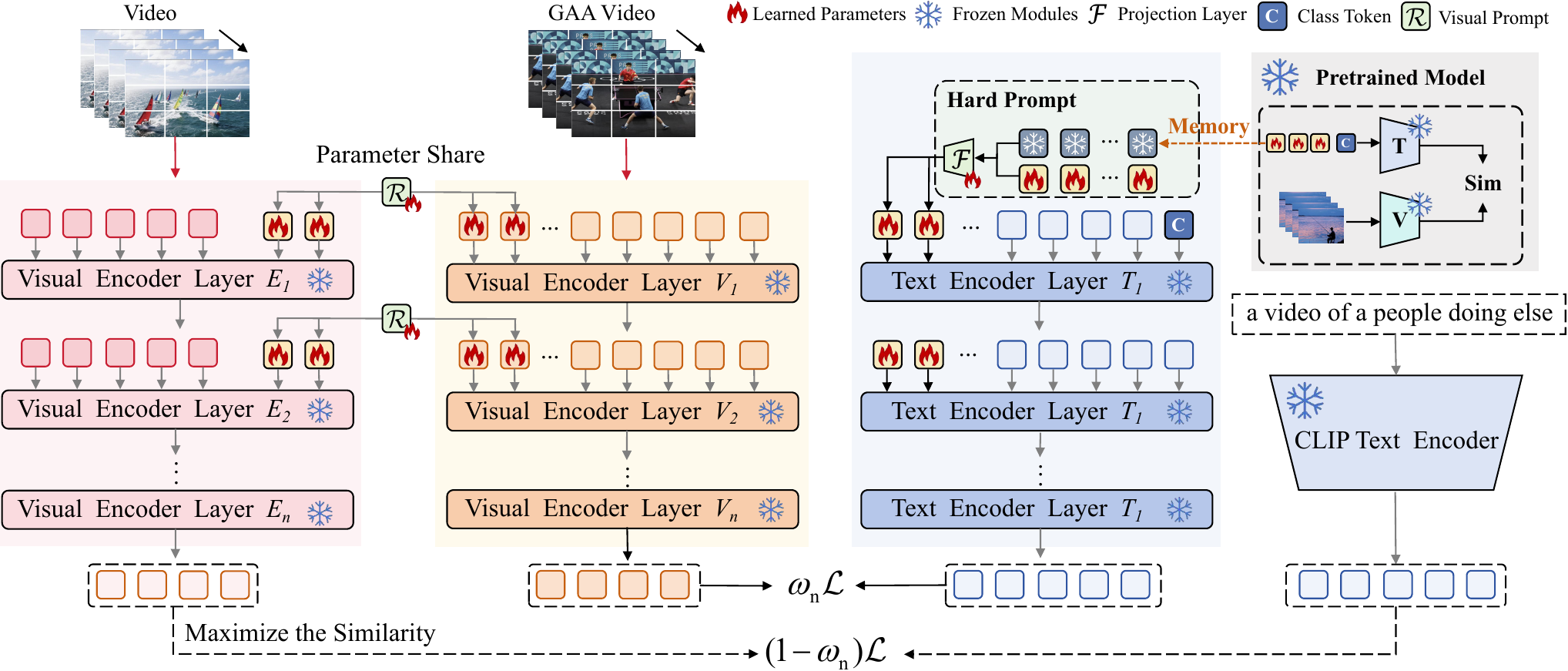}
    \caption{Architecture of the Generic Attribute Anchors CLIP (GA\textsuperscript{2}-CLIP) method for multi-modal prompt learning. The approach optimizes the model by adapting the vision and language branches, where only input prompts are learned while keeping the remainder of the model frozen.}
    \label{fig3}    
\end{figure*}

Our method aims to transfer the capabilities of CLIP to video action recognition tasks through visual language modeling and visual prompt learning. As shown in Fig.~\ref{fig2}(a), existing video text based approaches use the classical paradigm of soft prompt tokens with hard categorical tokens passed as inputs to a text encoder, but this can easily lead to overfitting, degrading the generalization ability. Although utilizing the vanilla prompt constraints from CLIP improves the situation, as shown in Fig.~\ref{fig2}(b), it also reduces the prompt learning capability. To address this problem, this paper proposes a video prompt learning method called GA\textsuperscript{2}-CLIP, which utilizes generic attribute anchors instead of the vanilla prompts of CLIP to constrain the prompt learning process, as shown in Fig.~\ref{fig2}(c). With generic attribute anchors, soft prompts are able to reduce spurious correlations with complex background information in the video, thus improving the video recognition performance of the video language model in unknown categories. In addition, to further improve the generalization ability of textual prompts, we introduce a pre-trained hard prompt for limiting the overfitting of learnable textual prompts in the base category task. For each video task, we can share the same pre-trained hard prompt as a supervisor and integrate it into our GA\textsuperscript{2}-CLIP for model fine-tuning, the overall is shown in Fig.~\ref{fig3}. 

\subsection{Visual and Text Encoding}

\noindent\textbf{Visual Encoding:} For the input video $\smash{\mathcal{V} \in \mathbb{R}^{3\times T\times H\times W}}$ with $T$ frames and $H\times W$ spatial size, the $t$-th frame $\{I_t\}_{t=1}^{T}$ is divided into $\smash{\{\mathbf{x}_{k,t}\}_{k=1}^{M} \in \mathbb{R}^{3\times P^2}}$ patch sizes according to the ViT architecture, $k$ denotes the patch number and $M=H\times W/P^2$. All patches are projected through a linear layer $p$ into a set of patch embedding vectors $\smash{\mathcal{E}_{0,t}\in \mathbb{R}^{M\times d_v}}$, $d_v$ being the dimension of the token and for the i-layer transformer $L$ can be represented as $\smash{\mathcal{E}_{i,t}}$. After adding a learnable class token $\mathbf{C}_{i,t}$ and positional encoding $\mathbf{e_0}$ , the frame-level encoding $E_v$ of the video can be expressed mathematically as:
\begin{equation}
    \mathcal{E}_{0,t}=[\mathbf{C}_{0,t},p^\mathbf{T} \mathbf{x}_{k,t},...,p^\mathbf{T} \mathbf{x}_{M,t}]+\mathbf{e_0},
\end{equation}
\begin{equation}
    E_v=[\mathcal{E}_{i,t}]=L_i([\mathcal{E}_{i-1,t}]), i=1,2,...,l,
\end{equation}
where $l$ is the number of transformer layers. To obtain the final video feature representation, we obtain the class token of the output of the last layer of the transformer and map it to a common V-L embedding space through the projection layer.

\noindent\textbf{Text Encoding:} The pre-trained text encoder is a 12-layer transformer structure with an embedding dimension of 512 and a context length of 77, which employs a Multi-Headed Self-Attention Mechanism (MHSA) backed feed-forward network (FFN) to capture interword associations and contextual information in the text sequence. Vanilla CLIP uses a fixed template of manually designed text prompts, e.g., “\texttt{a video of a person doing [CLS]}”, and subsequently uses a text encoder $E_t$ to obtain representation vectors for each category.

\noindent\textbf{Prompt Learning:} Unlike fixed prompts that require hand-crafted image-text alignment, the prompt learning approach quickly adapts to the downstream task by learning suitable soft prompts. Specifically, $M$ learnable soft tokens are utilized to replace the originally fixed textual prompts after connecting with the classification token \texttt{CLS} as new input to a text encoder. The form is as follows:
\begin{equation}
    W_P = [P_1][P_2]...[P_M][CLS],
\end{equation}
where $M$ denotes the length of the soft prompt token. In addition to adding soft prompt markers at the text input, existing approaches explore ways to introduce soft prompt token at deeper layers. For the text encoder $E_t$, this is implemented by re-adding soft prompt tokens in each layer of the Transformer and removing them after the computation of self-attention in each layer is completed, repeating the above process for each layer. The process for the i-th layer can be represented as:
\begin{equation}
    E_t= L_i([P_{i-1}, CLS_{i-1}]),i=1,2,...,l,
\end{equation}
where $L_i$ represents the i-th transformer block and $P_i$ denotes the set of learnable soft prompt tokens, defined as $P_i ={[P_1]_i, ..., [P_M]_i}$. For the visual encoder, the soft prompt addition process can still be followed for the text encoder by adding N soft prompt tokens in the first layer of the visual transformer, and similarly re-adding them in the deeper layers and deleting them after the completion of the self-attention computation. Formally this can be expressed as:
\begin{equation}
    E_v= L_i([\mathcal{E}_{i-1,t}, Q_{i-1}]),i=1,2,...,l,
\end{equation}
where $Q_i$ represents the set of learnable soft prompt tokens in visual transformer, defined as $Q_i ={[Q_1]_i, ..., [Q_N]_i}$, $N$ is the length of the visual soft prompt token.

\subsection{Pre-training Hard Prompt Token}

\noindent\textbf{Pre-training Stage:} We use a two-step tuning strategy to implement our pre-trained hard prompt tokens: first, we choose a small dataset that can better align video and text, which does not need to be very large but as informative as possible, and moderately fine-tune the original prompt vectors in the tuned main model in order to obtain the pre-trained prompts $P'$, which is subsequently frozen continuously during training.

While frozen tuning prompts $P'$ can ensure generalization ability during new category inference, which works with trainable tuning prompts to decouple the base category and the new category tasks \cite{li2025dpc}, our work explores the impact of this ability on video text alignment. Previous work embedded attribute hard prompts into text as they could get rich enough in inter-class semantics to achieve the expected gains. However, it is difficult to predefine effective semantic attributes for video, and the original purpose of the soft prompt token is to expect the model to learn effective semantic information, and although we cannot explain exactly what it learns, it turns out to be effective. We believe that the tuning prompt learns a generic projection, similar to a certain 'attribute' of a video, which is fixed by methods such as ATPrompt \cite{li2024atprompt} using VLMs, but is difficult to implement for video tasks. The approach in this paper is to make the model learn to summarize this 'attribute', and formally, the proposed hard prompt tokens belong to a special form of attribute prompts.

\noindent\textbf{Coupling of Double Prompts:}
In order to realize the coupling collaboration between the frozen prompts $P'$ and the soft prompts $P$, we need to maintain a dual-prompt input model during the fine-tuning and inference phases. Previous approaches use fixed weights to realize the coupling of inputs and re-decouple them when the novel and base category are tested, which is less flexible in video tasks. To address this problem, we use a set of nonlinear projection layers to realize the coupling of frozen prompts and soft prompts $\tilde{P}$, defined as:
\vspace{-0.1cm}
\begin{equation}
   \tilde{P}=Proj(P',P),
\end{equation}
where the nonlinear projection layer consists of a linear layer, the ReLu function and the Dropout layer. The introduction of nonlinear units can alleviate the problem of non-convergence caused by the uneven distribution between the randomly initialized soft prompt $P$ and the frozen hard prompt $P'$, allowing $\tilde{P}$ to learn more complex feature relationships. The above design avoids manual decoupling during base category and new category inference while ensuring model integrity, and enables flexible matching of potential features of freezing prompts $P'$ and tuning prompts $P$ across tasks. More coupling approaches will be explored in the Appendix.

\subsection{Generic Attribute Anchor}
\noindent\textbf{Motivation Analysis:} Previous work explores the negative impact caused by activating spurious correlations with background information, and in this section we rethink the impact mechanism of background attributes. Correctly distinguishing between foreground and background in image recognition tasks is one of the key factors affecting performance, and the introduction of negative background prompts (e.g., \texttt{the background of a [CLS]}) can effectively improve performance. However, this does not seem to be valid for video tasks, where the identification of many actions depends on certain background information. For example, water in the background plays a key role in distinguishing between volleyball and water volleyball, but cannot be relied upon for recognizing the volleyball action itself. Therefore, it is difficult to guide model learning by fixed attribute definitions.

From another point of view, the background information of videos is too complex, and when the data is few, it is easy to produce a large number of spurious correlations of background information, resulting in a narrowing of the semantics of the text that has been aligned with a large number of images and texts. Therefore, we expect to introduce a set of task-irrelevant videos as generic attribute anchors, which can make the textual prompts become “moderate” in the semantic training process.
As a result, the model is able to focus on the category names themselves to which the different videos refer, reducing the amount of spurious correlation of background semantics.

\noindent\textbf{Attribute Anchors and Prompts:}
For the creation of attribute anchors, our process is based on two core guidelines: 1) the background information of the selected videos is rich enough to be semantically aligned with the text, and 2) the original category labels cannot be the same as the labels of the videos to be trained. To this end, we design two independent steps to create attribute anchors: first, we randomly select a number of videos in irrelevant labels in a dataset with better video semantic alignment (e.g., Kinetics), and preprocess them using an independent sampler that follows the settings of the main backbone model. We then randomly feed these videos into the training pipeline using an iterator. To save resources, it can be fed to the video encoder uniformly with the main video in a concatenated manner and sliced for recovery at the output.

These videos can use a uniform prompt label called attribute anchor prompt. Attribute anchor prompts are used in a fixed way, e.g., \texttt{a video of a person doing something else}. All word inputs $V$ are fed to the text encoder with fixed prompts, a process denoted as:
\begin{equation}
    W_v = [V_1][V_2]...[V_N],
\end{equation}
where $N$ denotes the length of the attribute anchor prompts. It is noteworthy that attribute anchor prompts belong to a special pattern similar to category-directed prompts, which can be incorporated into the general prompting framework by replacing the last word with the class name.

\noindent\textbf{Training method:} During training, we jointly optimize textual and visual prompts, generic attribute anchors, and generic attribute prompts, constraining the model to learn the high similarity between action features in a video and the corresponding category embeddings, while reducing the similarity with other categories. Specifically, given an annotated video classification dataset containing $K$ category labels $C=\{ c_i \}^K_{i=1}$, for the input video $x$, firstly, feature extraction is performed by a visual coder to obtain the feature vector $f= E_v(x)$; meanwhile, a series of textual descriptions $t$ are generated for each category using a soft prompt template; subsequently these textual descriptions are fed into the text encoder to obtain the textual features $w= E_t(t)$. The standard comparison loss learning is:
\begin{equation}
   \mathcal{L}(y \mid x) = \frac{\exp(\cos(f, w_y)/\tau)}{\sum_{i=1}^N \exp(\cos(f, w_i)/\tau)},
\end{equation}
where $\tau$ is the temperature parameter and cos(·, ·) denotes
cosine similarity, $N$ is the number of seen classes. 

For the generic attribute anchors input sample $x^n$, the visual coder is used to obtain the vector $g=E_v(x^n)$. We define the embedding vector of its attribute prompts as $n_c=\{ v_1,v_1,...,v_\mathcal{M} \}$, where $v_i$ is the word vector of the attribute prompt and $\mathcal{M}$ represents the number of prompt words, next let $C=\{ c_j^n \}^K_{
j=1}$. At this point the comparison between generic attribute anchors and generic attribute prompts can be represented as:
\begin{equation}
   \mathcal{L}_n(y \mid x^n) = \frac{\exp(\cos(g, w_y^n)/\tau)}{\sum_{j=1}^K \exp(\cos(g, w_j^n)/\tau)},
\end{equation}
where $K$ is the number of anchor attribute prompts. The function is optimized using cross-entropy. Finally, we introduce a specific weighting factor $\omega_n$ to redistribute the two losses to obtain the final total loss $\mathcal{L}_s$:
\begin{equation}
   \mathcal{L}_s=\omega_n\mathcal{L}+(1-\omega_n)\mathcal{L}_n,
\end{equation}
the above design allows the model to flexibly adjust the weight coefficients based on the number of training datasets and generic attribute anchor datasets without the need to specifically adjust the number of  attribute anchor videos. 

\section{Experiments}
\noindent\subsection{Implementation Details}

\begin{table*}[h]
\centering

\resizebox{\textwidth}{!}
{
\begin{tabular}{lccc|ccc|ccc|ccc}
\toprule
\multirow{2}{*}{Method} & 
\multicolumn{3}{c}{HMDB-51} & 
\multicolumn{3}{c}{UCF-101} & 
\multicolumn{3}{c}{SSv2} & 
\multicolumn{3}{c}{Kinetics-400} \\
\cmidrule(lr){2-13}
& Base & Novel & HM & Base & Novel & HM & Base & Novel & HM & Base & Novel &HM \\
\midrule
\multicolumn{13}{c}{Adapting pre-trained image VL models} \\
\midrule
Vanilla CLIP \cite{radford2021learning} & 53.3 & 46.8 & 49.8 & 78.5 & 63.6 & 70.3 & 4.9 & 5.3 & 5.1 & 62.3 & 53.4 & 57.5 \\
ActionCLIP \cite{wang2023actionclip} & 69.1 & 37.3 & 48.5 & 90.1 & 58.1 & 70.7 & 13.3 & 10.1 & 11.5 & 61.0 & 46.2 & 52.6 \\
X-CLIP \cite{ni2022expanding} & 69.4 & 45.5 & 55.0 & 89.9 & 58.9 & 71.2 & 8.5 & 6.6 & 7.4 & 74.1 & 56.4 & 64.0 \\
A5 \cite{ju2022prompting} & 46.2 & 16.0 & 23.8 & 90.5 & 40.4 & 55.8 & 8.3 & 5.3 & 6.4 & 69.7 & 37.6 & 48.8 \\
\midrule
\multicolumn{13}{c}{Tuning pre-trained image VL models} \\
\midrule
CLIP image-FT \cite{rasheed2023fine} & 62.6 & 47.5 & 54.0 & 86.4 & 65.3 & 74.4 & 9.2 & 8.5 & 8.8 & 72.9 & 58.0 & 64.6 \\
CLIP text-FT \cite{rasheed2023fine} & 70.0 & 51.2 & 59.1 & 90.9 & 67.4 & 78.3 & 12.4 & 9.5 & 10.8 & 73.4 & 59.7 & 65.8 \\
ViFi-CLIP \cite{rasheed2023fine} & 73.8 & 53.3 & 61.9 & 92.9 & 67.7 & 78.3 & 16.2 & 12.1 & 13.9 & \underline{76.4} & \underline{61.1} & \underline{67.9} \\

VTD-CLIP \cite{zhu2026vtd} & \textbf{78.4} & \textbf{63.5} & \textbf{70.0} & \textbf{95.5} & \textbf{73.7} & \textbf{83.2} & \textbf{17.8} & \textbf{13.9} & \textbf{15.4} & \textbf{78.5} & \textbf{63.5} & \textbf{70.1} \\
\midrule
\multicolumn{13}{c}{Prompt tuning pre-trained image VL models} \\
\midrule
ViFi-CLIP \cite{rasheed2023fine} & 77.1 & 54.9 & 64.1 & 95.1 & 74.1 & 83.6 & 15.8 & 11.5 & 13.3 & -- & -- & -- \\

ViLT-CLIP \cite{wang2024vilt} & 76.7 & 57.5 & 65.7 & 95.2 & 70.5 & 81.0 & 17.3 & 12.8 & 14.7 & \textbf{77.4} & 63.0 & 69.5 \\

\textbf{Ours} & \textbf{78.3} & \textbf{58.9} & \textbf{67.2} & \textbf{96.8} & \textbf{75.2}  & \textbf{84.6} & \textbf{18.7} & \textbf{14.3} & \textbf{16.2} & 77.0 & \textbf{63.3} & \textbf{69.5} \\
 & +1.2 & +4.0 & +3.1 & +1.7 & +1.1 & +1.0 & +2.9 & +2.8 & +2.9 & +0.6 & +2.2 & +1.6 \\
\bottomrule
\end{tabular}
}
\caption{Base-to-novel generalization ability performance: Comparison with other state-of-the-art methods on HMDB-51, UCF-101, SSv2 and Kinetics-400 datasets. Where Base refers to half of the video categories randomly selected for training and Novel consists of the remaining categories, HM refers to the harmonic mean, which is used to measure the trade-off between base accuracy and novelty accuracy. ViFi-CLIP is used as a baseline for comparison, and the gains are highlighted in the last row.}
\vspace{-0.2cm}
\label{tab1}
\end{table*}

\begin{table}[ht]
\centering

\resizebox{\linewidth}{!}
{
\begin{tabular}{lcccc}

\toprule
Method & Frames & Views &Top-1 & Top-5 \\
\midrule
\multicolumn{5}{c}{Uni-modal architectures} \\
\midrule
Uniformer-B (2023) & 32 & $4 \times 3$ & 83.0 & 95.4  \\
TimeSformer-L (2021) & 96 & $1 \times 3$ & 80.7 & 94.7  \\
Swin-L (2022) & 32 & $4 \times 3$ & 83.1 & 95.9  \\
\midrule
\multicolumn{5}{c}{Adapting pre-trained image VL models} \\
\midrule
ActionCLIP (2021) & 32 & $10 \times 3$ & 83.8 & 96.2  \\
X-CLIP (2022) & 16 & $4 \times 3$ & \textbf{84.7} & \textbf{96.8}  \\
A6 (2022) & 16 & -- & 76.9 & 93.5 \\
\midrule
\multicolumn{5}{c}{Prompting pre-trained image VL models} \\
\midrule
ViLT-CLIP B/16 (2024)& 16 & $4 \times 3$ & 77.6 & 94.5 \\
Ours & 16 & $4 \times 3$ & \textbf{77.8} & \textbf{95.1}\\
\bottomrule
\end{tabular}
}

\caption{Fully-supervised action recognition performance comparison on Kinetics-400.}
\vspace{-0.5cm}
\label{tab2}

\end{table}

We use five benchmark datasets in our experiments: the Kinetics-400 and Kinetics-600 \cite{kay2017kinetics}, HMDB-51 \cite{kuehne2011hmdb}, UCF-101 \cite{soomro2012ucf101}, and Something-Something v2 (SSv2) \cite{goyal2017something}.

\noindent\textbf{Kinetics-400 and Kinetics-600 datasets:} the Kinetics-400 dataset contains 400 human action categories, consisting of video clips from different YouTube platforms, each of which is about 10 seconds long. The dataset contains about 240,000 training videos and 20,000 validation videos. Kinetics-600 is an extension of Kinetics-400 and contains about 650,000 video clips covering 600 action categories, of which about 410,000 are training videos and 29,000 are validation videos.

\noindent\textbf{HMDB-51:} The HMDB-51 dataset contains 6,766 short videos from 51 categories. These video clips are collected from movies, the web, and public databases, and cover a wide range of complex scenes and viewpoints. The dataset is divided into three separate training and validation sets: 3,783 segments for training and 1,530 segments for validation.

\noindent\textbf{UCF-101:} The UCF-101 dataset contains 13,320 short video clips in 101 categories from the YouTube website, covering a variety of human activities, musical instrument playing, sports, and other scenarios. The standard training set of this dataset contains 9,537 videos and the evaluation set contains 3,783 videos which are divided into three subsets for testing.

\noindent\textbf{Something-Something v2 (SSv2):} The SSv2 dataset is a collection of video clips containing a large number of humans performing actions with everyday objects, covering 174 action classes. The dataset is primarily used to train models that are capable of fine-grained understanding of human actions, such as the action of placing an item at a specific location or on another object. The standard division contains 168,913 training videos and 24,777 validation videos.

\noindent\textbf{Evaluation Protocols:}
For the \textbf{few-shot experiments}, we follow the generalized K-shot split scheme, where K-shot denotes that K samples from each category are randomly selected for training. Specifically, 2, 4, 8 and 16 samples are used for training for the three datasets HMDB-51, UCF-101 and SSv2, respectively. The training and validation process only considers the first split of the HMDB-51 and UCF-101 datasets, while Kinetics and SSv2 are evaluated with the model on the full validation set. The setup uses 32 frames and evaluates with single-view inference. For \textbf{base-to-new generalization experiments}, these splits evenly divided the total categories into two equal groups: the most frequent categories as base categories and the less frequent categories as new categories, with 16 random samples of action categories per category. Models are evaluated on the first split of the HMDB-51 and UCF-101 datasets as well as on the full validation splits of Kinetics and SSv2. Both the base-to-new and few-sample experiments used 32 frames and evaluates with single-view inference.

\noindent\textbf{Implementation Details:} In the GA\textsuperscript{2}-CLIP experiments, all randomly sampled frames are preprocessed to a spatial size of 224 × 224. The maximum number of text characters is limited to 77 according to the CLIP \cite{clip} settings. We use the AdamW optimizer and set the weight decay factor to 0.001. We carry out the analysis in four experimental scenarios: zero-shot, base-to-novelty generalization, few-shot, and fully supervised. In the base-to-novelty generalization scenario, we train 12 epochs with a batch size of 64 and a learning rate of 4e-2 using 16 vision and language prompts in the first 9 layers of the CLIP vision and language encoders, respectively. In the action recognition task in few-shot scenarios, we optimize the GA\textsuperscript{2}-CLIP model by using 10 visual and language prompt in all transformer layers. The model is trained for 30 epochs using a cosine decay scheduler with the initial learning rate set to 8e-3. 
In fully supervised scenarios, we train GA\textsuperscript{2}-CLIP for 30 epochs on the Kinetics-400 dataset using a batch size of 256 and a learning rate of 4e-2.

\begin{table*}[ht]
\centering

\resizebox{\textwidth}{!}
{
\begin{tabular}{lcccc|cccc|cccc}
\toprule
\multirow{2}{*}{Method} & 
\multicolumn{4}{c}{HMDB-51} & 
\multicolumn{4}{c}{UCF-101} & 
\multicolumn{4}{c}{SSv2}
 \\
\cmidrule(lr){2-13}
& \textit{K}=2 & \textit{K}=4 & \textit{K}=8 & \textit{K}=16 & \textit{K}=2 & \textit{K}=4 & \textit{K}=8 & \textit{K}=16 & \textit{K}=2 & \textit{K}=4 & \textit{K}=8 & \textit{K}=16  \\
\midrule
\multicolumn{13}{c}{Adapting pre-trained image VL models} \\
\midrule
Vanilla CLIP \cite{radford2021learning} & 41.9 & 41.9 & 41.9 & 41.9 & 63.6 & 63.6 & 63.6 & 63.6 & 2.7  &2.7 & 2.7 & 2.7 \\
ActionCLIP \cite{wang2023actionclip} & 47.5 & 57.9 & 57.3 & 59.1 & 70.6 & 71.5 & 73.0 & 91.4 & 4.1 & 5.8 & 8.4 & 11.1 \\
X-CLIP \cite{ni2022expanding} & 53.0 & 57.3 & 62.8 & 64.0 & 48.5 & 75.6 & 83.7 & 91.4 & 3.9 & 4.5 & 6.8 & 10.0 \\
A5 \cite{ju2022prompting} & 39.7 & 50.7 & 56.0 & 62.4 & 71.4 & 79.9 & 85.7 & 89.9 & 4.4 & 5.1 & 6.1 & 9.7 \\
\midrule
\multicolumn{13}{c}{Tuning pre-trained image VL models} \\
\midrule
CLIP image-FT \cite{rasheed2023fine} & 49.6 & 54.9 & 57.8 & 62.0 & 74.4 & 79.1 & 85.3 & 90.5 & 4.9 & 6.0 & 7.2 & 10.4 \\
CLIP text-FT \cite{rasheed2023fine} & 54.5 & 61.6 & 63.1 & 65.0 & 80.1 & 82.8 & 85.8 & 88.1 & 6.2 & 6.1 & 6.3 & 9.1 \\
ViFi-CLIP \cite{rasheed2023fine} & 57.2 & 62.7 & 64.5 & 66.8 & 80.7 & 85.1 & 90.0 & 92.7 & 6.2 & 7.4 & 8.5 & \textbf{12.4} \\
OST \cite{chen2024ost} & \textbf{59.1} & \textbf{62.9} & \textbf{64.9} & \textbf{68.2} & \textbf{82.5} & \textbf{87.5} & \textbf{91.7} & \textbf{93.9} & \textbf{7.0} & \textbf{7.7} & \textbf{8.9} & 12.2 \\
VLPA-CLIP \cite{wang2025vlpa} & 54.7  & 59.4 & 63.7 & 66.6 & 78.2 & 85.0 & 89.5 & 92.7 & 5.8 & 6.4 & 8.6 & 11.4 \\
\midrule
\multicolumn{13}{c}{Prompt tuning pre-trained image VL models} \\
\midrule
ViFi-CLIP \cite{rasheed2023fine} & 57.8 & 60.0 & 64.8 & 67.4 & 84.0 & 87.1 & 90.4 & 93.1 & 7.3 & 8.1 & 9.8 & 13.0 \\

ViLT-CLIP \cite{wang2024vilt} & 60.6 & 61.9 & 66.9 & 69.6 & 85.3 & 90.0 & 91.3 & 93.8 & \textbf{7.8} & 9.4 & 10.3 & 13.2 \\
\textbf{Ours} & \textbf{61.9} & \textbf{64.3} & \textbf{68.1} & \textbf{70.8} & \textbf{90.0}  & \textbf{91.8} & \textbf{93.8} & \textbf{95.5} & 6.8 & \textbf{9.9} & \textbf{10.6} & \textbf{14.7} \\
 & +4.1 & +4.3 & +3.3 & +3.4 & +6.0 & +4.7 & +3.4 & +2.4 & -0.5 & +1.8 & +0.8 & +1.7 \\
\bottomrule
\end{tabular}
}
\caption{Few-shot setting performance: Comparison with other state-of-the-art methods on HMDB-51, UCF-101 and SSv2 datasets. We performed few-shot experiments using 2, 4, 8, and 16 video sequences in different datasets, with a different number of video sequences for each category. ViFi-CLIP is used as a baseline for comparison, and the gains are highlighted in the last row.}
\vspace{-0.2cm}
\label{tab3}
\end{table*}

\begin{table}[h]
\centering
\begin{tabular}{lcc}
\toprule
Method & HMDB-51 & UCF-101 \\
\midrule
\multicolumn{3}{c}{Uni-modal zero-shot action recognition models} \\
\midrule
ZSECOC \cite{qin2017zero} & 22.6 $\pm$ 1.2 & 15.1 $\pm$ 1.7 \\
UR \cite{zhu2018towards} & 24.4 $\pm$ 1.6 & 17.5 $\pm$ 1.6 \\
E2E \cite{brattoli2020rethinking} & 32.7 & 48.0 \\
ER-ZSAR \cite{chen2021elaborative} & 35.3 $\pm$ 4.6 & 51.8 $\pm$ 2.9 \\
\midrule
\multicolumn{3}{c}{Adapting pre-trained image VL models} \\
\midrule
Vanilla CLIP \cite{radford2021learning} & 40.8 $\pm$ 0.3 & 63.2 $\pm$ 0.2 \\
ActionCLIP \cite{wang2023actionclip} & 40.8 $\pm$ 5.4 & 58.3 $\pm$ 3.4 \\
X-CLIP \cite{ni2022expanding} & 44.6 $\pm$ 5.2 & 72.0 $\pm$ 2.3 \\
A5 \cite{ju2022prompting} & 44.3 $\pm$ 2.2 & 69.3 $\pm$ 4.2 \\
\midrule
\multicolumn{3}{c}{Prompting pre-trained image VL models} \\
\midrule
ViLT-CLIP B/16 \cite{wang2024vilt} & 45.3 $\pm$ 0.9 & \textbf{73.6} $\pm$ \textbf{1.1} \\
MVPC-CLIP B/16 \cite{zhan2026mvpc} & 47.7  & 74.5  \\
Ours  & \textbf{48.3} $\pm$ \textbf{0.5} & 68.2 $\pm$ 1.2 \\
\bottomrule
\end{tabular}
\caption{Zero-shot action recognition performance on HMDB-51 and UCF-101 datasets. Use 32 frames and evaluate with a single-view inference. }
\label{tab4}
\end{table}

\noindent\subsection{Cross-Domain Generalization and Adaptation}
We evaluate the cross-domain generalization and adaptation capabilities of the Generic Attribute Anchor tuning method GA\textsuperscript{2}-CLIP in a video action recognition task with the following experimental setups: 1) base-to-novel setting, 2) full-supervised setting, 3) few-shot setting, 4) zero-shot setting.

\noindent\textbf{Base-to-New Generalization.} In Tab.~\ref{tab1}, we evaluate the performance of ViLT-CLIP on the four datasets, HMDB-51, UCF-101, SSv2, and K-400, in terms of its ability to generalize from base to novel categories. Compared with the baseline method ViFi-CLIP, the proposed method gains a significant improvement in both base and novel category accuracy. The model achieves a better balance between base and novel category accuracy, resulting in the best harmonic mean across all datasets. Notably, GA\textsuperscript{2}-CLIP in the SSv2 dataset, which has relatively low background effects, shows unexpected results, with +2.9\% and +1.5\% improvement compared to the prompt fine-tuning baseline ViFi-CLIP and the state-of-the-art ViLT-CLIP. The experimental results indicate that the proposed method can effectively improve the generalization performance of the video task.

\noindent\textbf{Fully-Supervised:} We compare the action recognition performance of GA\textsuperscript{2}-CLIP trained on the Kinetics-400 dataset with other unimodal and prompt methods in Tab.~\ref{tab2}. Although the proposed method focuses on cross-domain adaptation and generalization capabilities, it still leads the state-of-the-art ViLT-CLIP in prompt fine-tuning methods with a Top-1 accuracy +0.2\%. There is a gap with other unimodal and VL pre-training methods, but our method only needs to train learnable prompts so it is more flexible.

\noindent\textbf{Few-shot Learning:} To evaluate the effectiveness and generalization ability of the proposed method under limited data conditions, we show the performance of GA\textsuperscript{2}-CLIP in few-shot scenario, as shown in Tab.~\ref{tab3}. As the number of videos increases, the accuracy improves significantly. Intuitively, we achieve performance gains on all three datasets compared to the previous state-of-the-art method. This indicates that GA\textsuperscript{2}-CLIP not only enhances the model generalization ability, but also has strong resistance to overfitting.

\noindent\textbf{Zero-shot Video Recognition:} In the zero-shot setting, we train our model on Kinetics-400, a large-scale video action recognition dataset, and evaluate on two validation sets: HMDB-51 and UCF-101. As shown in Tab.~\ref{tab4}, compared to the previous state-of-the-art, our approach achieves a performance gain of +3.0\% on HMDB-51, while exhibiting marginal degradation on UCF-101. It should be noted that the full K400 dataset is used for training here. It is challenging to design a generic attribute anchor that can adapt to such a large-scale dataset, resulting in suboptimal performance. Nevertheless, it still outperforms most state-of-the-art methods. In summary, GA\textsuperscript{2}-CLIP demonstrates strong cross-domain generalization capabilities in video action recognition tasks.


\subsection{Pre-trained Hard Prompt}

\noindent\textbf{Component Validity:} In this section, we decoupled hard prompt tuning and generic attribute anchors, testing the effectiveness of different components, the results are shown in Tab.~\ref{tabf1}. The results demonstrate that HPT effectively enhances the performance of base classes, while GAA shows more pronounced improvements for novel classes. The best results are achieved when both components are used together. The experimental findings align with the expected objectives, validating the effectiveness of the approach.

\begin{table}[h]
\centering

\begin{tabular}{c|ccc}
\toprule
Component &  Base & Novel & HM  \\
\midrule
Baseline &  73.8 & 53.3 & 61.9 \\
+ HPT&  77.6(+3.8) & 55.3(+2.0) & 64.6(+2.7)  \\
+ GAA &  75.7(+1.9) & 57.9(+4.6) & 65.6(+3.7)  \\
\midrule
Full &  78.3(+4.5) & 58.9(+5.6) & 67.2(+5.3)  \\
\bottomrule
\end{tabular}
\caption{Validity of Different Components. HPT denotes Hard Prompt Tuning and GAA denotes Generic Attribute Anchors.}
\label{tabf1}
\end{table}

\noindent\textbf{Hard Prompt Training Source:} In this section we study the impact of different hard prompt data sources on performance. In the main experiment, the hard prompts for HMDB-51, UCF-101 and SSv2 are derived from the Kinetics-400 base category training set in order to avoid the suspicion of data leakage. For Kinetics-400, the hard prompts for the main experiment are derived from a training set consisting of an additional 100 classes of videos unrelated to the Kinetics-400 labels. 


\begin{table}[h]
\centering

\begin{tabular}{c|ccc|c}
\toprule
Source &  Base & Novel & HM & $\Delta$ \\
\midrule
- &  76.4 & 61.1 & 67.9 & \\
UCF-101 &  76.9 & 62.2 & 68.8 & +0.8 \\
HMDB-51 &  76.6 & 61.4 &  68.2 & +0.3 \\
SSv2 &  76.3 & 58.7 & 66.4 & -1.5 \\
\midrule
K-600 &  \textbf{77.0} & \textbf{63.3} & \textbf{69.5} & +1.6 \\
\bottomrule
\end{tabular}
\caption{Impact of different hard prompt data sources on Kinetics-400 base to novel generalization performance.}
\label{tab5}
\end{table}

In addition, we also study the effect of different hard prompt data sources on the performance of base to novel generalization, and the results are shown in Tab.~\ref{tab5}. In our experiments, we only use base category from different datasets for pre-training (K-600 uses independent 100 class-independent videos), and the results show that the performance gains a certain amount of improvement except for SSv2, and the improvement using K-600 is more significant +1.6\%. SSv2 focuses more on temporal related actions and ignores the background semantic correlation, thus producing a negative effect.

\begin{figure}[ht]
    \centering
    \includegraphics[width=0.8\linewidth]{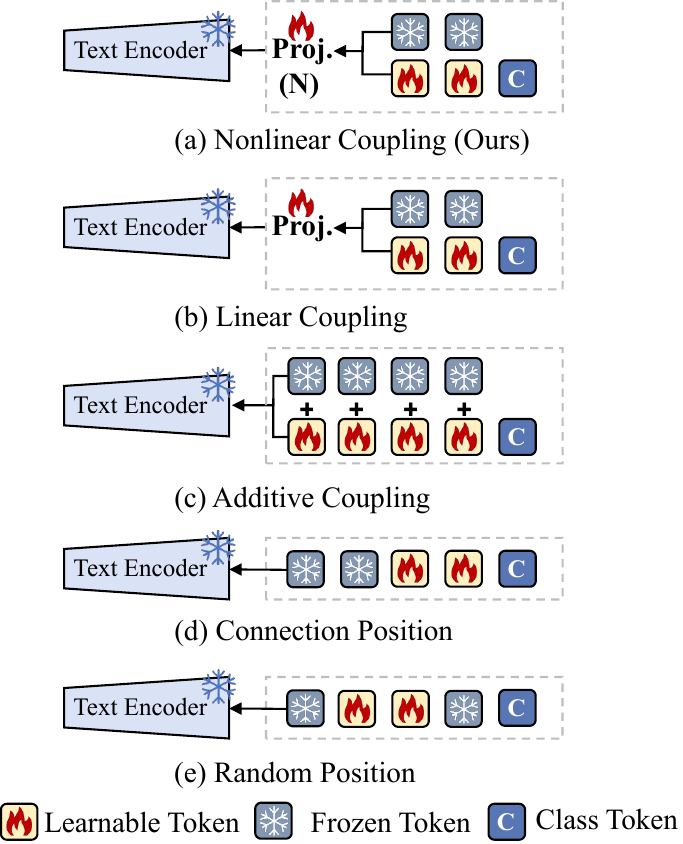}
    \caption{Comparison of different hard and soft prompt token coupling methods.}
    \label{fig4}    
\end{figure}

\begin{table}[hb]
\centering

\begin{tabular}{c|ccc}
\toprule
Style &  Base & Novel & HM  \\
\midrule
Soft Prompt (Baseline) &  76.6 & 61.9 & 68.5 \\
\midrule
Nonlinear Coupling (Ours) &  77.0 & 63.3 & 69.5  \\
Linear Coupling &  76.9 & 62.2 &   68.7 \\
Additive Coupling &  76.5 & 59.6 &  67.0  \\
Connection Position &  76.4 & 60.1 & 67.2  \\
Random Position &  75.3 & 54.9 &  63.5\\
\bottomrule
\end{tabular}
\caption{Performance of different soft and hard prompt coupling methods.}
\label{tab6}
\end{table}

\noindent\textbf{Coupling Method:} We also investigate the effect of the way hard prompt tokens are coupled with soft prompt tokens. Fig.~\ref{fig4} illustrates the different ways of setting up, and Tab.~\ref{tab6} presents the different results. We find that using nonlinear coupling gives the best performance, and followed by linear coupling. All other coupling methods that directly interact with soft prompt tokens break the semantic representation.

\begin{figure*}[th]
    \centering
    \includegraphics[width=1.0\linewidth]{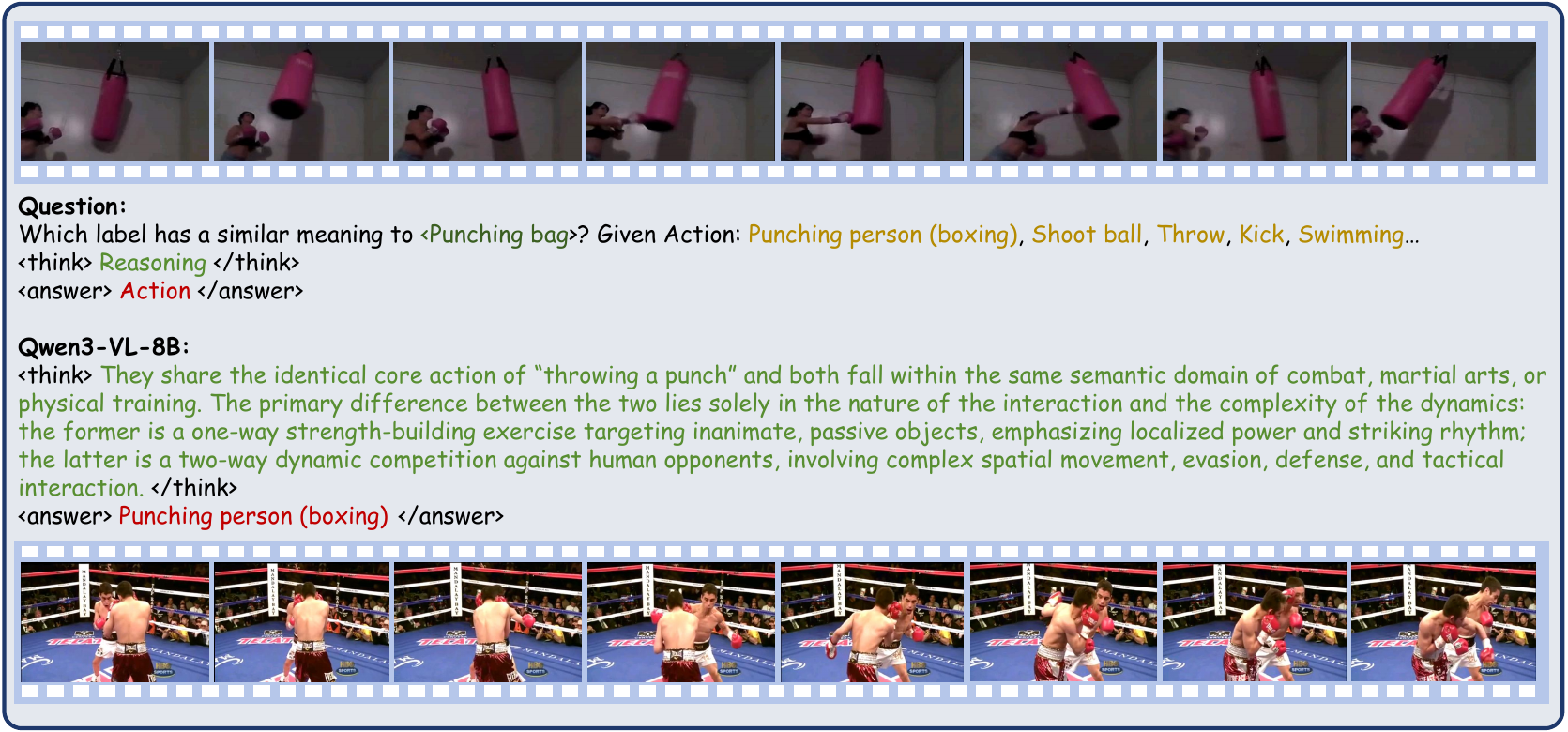}
    \caption{Use Qwen3-VL-8B \cite{yang2025qwen3} to generate matching results and descriptions for HMDB-51 and K-400 labels. Qwen3-VL-8B is capable of identifying actions with similar semantic meanings, enabling GA\textsuperscript{2}-CLIP to implicitly learn general semantics.}
    \label{fig8}    
\end{figure*}

\begin{table*}[ht]
\centering

\begin{tabular}{c|ccc}

\toprule
Action Pairs (Base vs. Novel)&	ViFi-CLIP&	GA\textsuperscript{2}-CLIP (Ours)	&TACS \\
\midrule
biking through snow/riding a bike &	81.7 / 56.5&	83.5 / 65.1	&snow (0.78 vs. 0.63) \\
ice climbing/rock climbing	&77.1 / 49.6&	78.1 / 63.7&	ice (0.52 vs. 0.43)\\
water skiing/skiing&	75.3 / 58.7&	77.4 / 60.1	&water (0.61 vs. 0.53)\\
diving cliff/springboard diving&	66.9 / 48.8&	69.3 / 57.4	&cliff (0.66 vs. 0.49)\\

\bottomrule
\end{tabular}
\caption{Cosine similarity of text features between the base class and the novelty class.}
\label{tab9}
\end{table*}

\subsection{About Generic Attribute Anchor}

\begin{figure}[ht]
    \centering
    \includegraphics[width=1.0\linewidth]{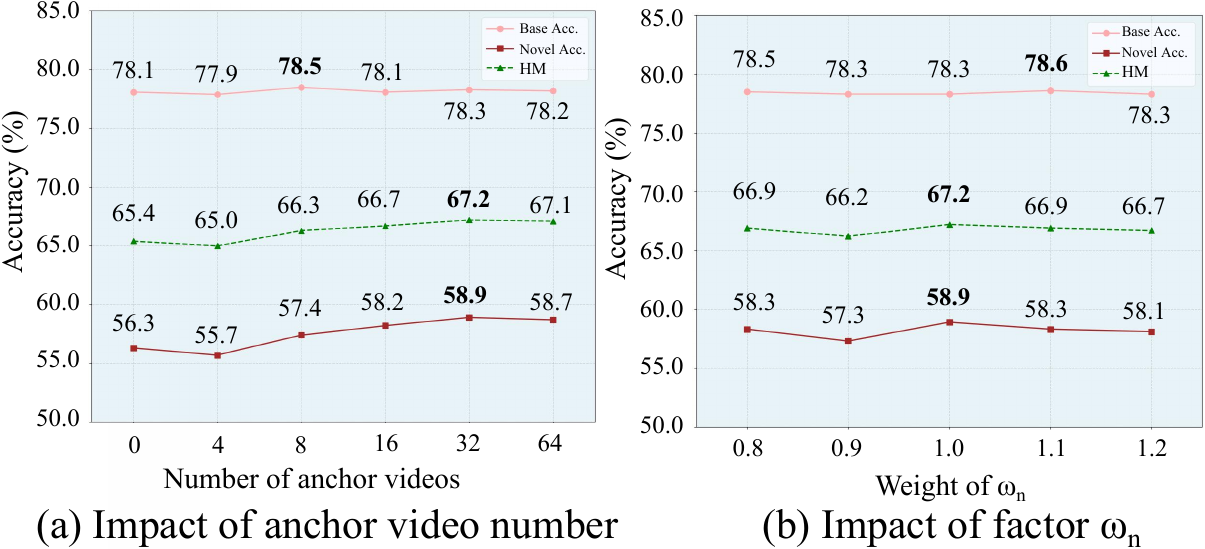}
    \caption{The effect of different settings on base to novel. (a) The effect of the number of different anchor videos, sampled from 4-64, with 0 indicating no use. (b) The effect of the fusion factor, here the weight of the vanilla factor is fixed to 1.0.}
    \label{fig5}    
\end{figure}

\noindent\textbf{Component analysis:} The generic attribute anchor videos used in the experiments are from the base category of K-400 (for K-400 , we use the novel category in K-600). Specifically, we manually or using an LLM model (e.g., Qwen \cite{achiam2023gpt, li2025tokenpacker,liu2024deepseek, lin2023video, yang2025qwen3}) filter out several labels among the labels of all anchor source datasets that may be semantically related but different from the training data labels (e.g., pushing cart vs. pushing; sit vs. read book; walking the dog vs. walk, etc.). Fig.~\ref{fig8} shows the matching results and the instructions for LLM generation for some of the HMDB-51 and K-400 labels. For labels with the same semantics, generic attribute anchors can serve to increase the samples; while similar labels can sufficiently avoid semantic narrowing. And then randomly sample $K$ videos from the selected labels to use as anchor videos, where the effect of the value of $K$ on the performance is shown in Fig.~\ref{fig5}(a). The results show that the performance gradually improves as the number of samples increases, which proves that generic attribute anchors are effective.

\begin{table}[h]
\centering

\begin{tabular}{c|ccc}
\toprule
Regularization &  Base & Novel & HM  \\
\midrule
Explicit regularization (Ours) &  78.3 & 58.9 & 67.2 \\
Margin Loss ($\lambda=0.1$) &  78.1 & 56.4 & 65.5  \\
Margin Loss ($\lambda=0.2$) &  77.7 & 55.1 & 64.5  \\

\bottomrule
\end{tabular}
\caption{Impact of different regularization methods on performance.}
\label{tab7}
\end{table}

In summary, the method of obtaining specific attributes \cite{li2024atprompt, tian2024argue, kim2024aapl, ding2024tree} can guide the model to “know what to learn”, but this is difficult for video tasks, especially those related to human movement. Our GA\textsuperscript{2}-CLIP serves as a semantic hub for these non-target attributes. By mapping diverse and task-irrelevant video content to a neutral text prompt, we explicitly model the semantic space of “everything else.” This contrast mechanism forces the model to learn text prompts that focus solely on specific, discriminative action primitives of the target class, rather than autonomously encoding irrelevant background attributes. It is termed an attribute anchor because it anchors the semantic baseline of the visual-language model’s representation space, thereby effectively preventing text embeddings from collapsing due to specific visual biases in the underlying training dataset.


\begin{figure*}[t]
    \centering
    \includegraphics[width=1.0\linewidth]{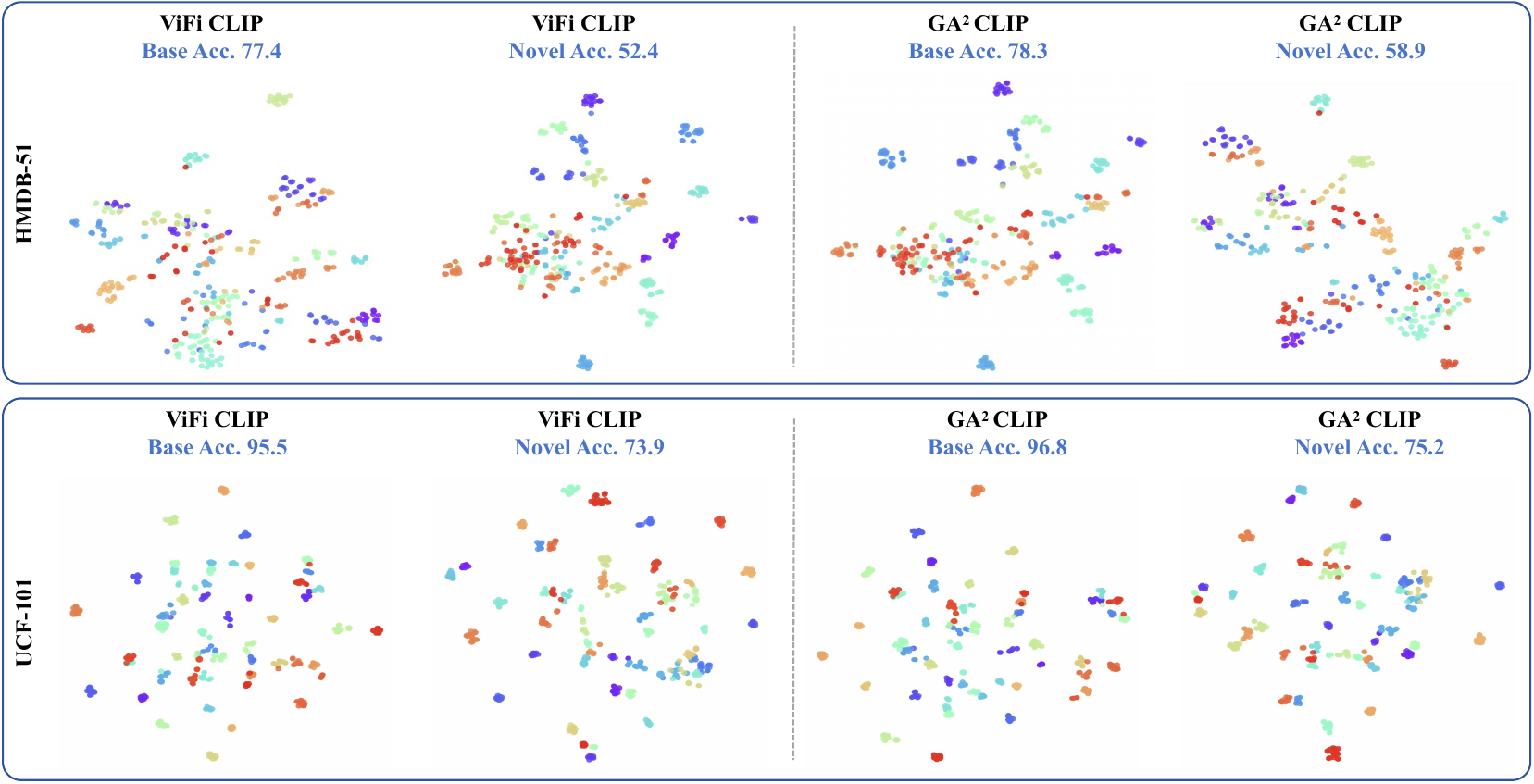}
    \caption{t-SNE visualizations for HMDB-51 and UCF-101 datasets. We perform t-SNE visualizations of the final visual embedded features of the base and novel category, and compare them with the baseline method ViFi-CLIP \cite{rasheed2023fine}, the results show that the proposed method has better separability.}
    \label{fig6}    
\end{figure*}

\noindent\textbf{Regularization settings:} In this subsection, we experiment with different regularization methods and parameter settings. Fig.~\ref{fig5}(b) shows the base to novel performance for different regularization factors, with the harmonic mean optimal at $\omega_n=1.0$. Tab.~\ref{tab7} shows the performance of the two different regularization methods and the results of utilizing direct regularization are better compared to the Margin Loss \cite{rosset2003margin} method.

\subsection{Visualization Related}
\label{sec:B}
Fig.~\ref{fig6} demonstrates the proposed method and the baseline approach ViFi-CLIP \cite{rasheed2023fine} on the HMDB-51 and UCF-101 datasets, base to novel experimental setup. We utilize t-SNE \cite{maaten2008visualizing} to visualize the embedding of the last layer of visual coder output, employing a maximum 500 sample limit. The results show that the fine-tuning prompt method enables linear dimensionality reduction with improved generalization performance in the presence of frozen other transformer layers. 

Further targeted ablation experiments, as shown in Tab.~\ref{tab9}, we separately analyzed the performance comparison of labels prone to semantic collapse in the B2N dataset under the two methods. Textual Action Cosine Similarity (TACS) represents the cosine similarity of features extracted from pure background vocabulary and pure action vocabulary, respectively, using a frozen text encoder. Taking the “ice climbing” label as an example, because the model overfits the background, the similarity between the fine-tuned text features “ice” and “climbing” in the ViFi method is significantly higher than that in GA\textsuperscript{2}-CLIP. This indicates that feature space collapse has occurred, and ‘climbing’ has been completely “ice-ified.” GA\textsuperscript{2}-CLIP forces the model to exclude general background information, demonstrating that action features are successfully retained and background dependencies are decoupled. When visual features from the novel “rock climbing” category are input, similarity matching naturally succeeds because the text side retains the pure semantic meaning of “climbing.

\subsection{Limitations and Future Works}
\label{sec:C}
In addition to the limitations discussed in the main paper, this work has the following issues that need to be addressed: 1) Although generic attribute anchors have achieved good results, they lack full explainability, especially when compared to methods such as attribute prompting. 2) No generic attribute anchor data source suitable for all video action recognition benchmarks has been found, but this work proves that the direction is promising and we will explore more video generic anchor methods in the future. 3) The temporal-oriented video task still lacks sufficient power, and traditional semantic correlation methods perform poorly in temporally relevant datasets such as SSv2, so we will explore more “action-oriented” attribute anchors in the future. In summary, future work may focus on exploring more effective semantic anchors for video using methods such as Multimodal Large Language Models (MLLMs).
\section{Conclusion}

In this work, we propose a video prompt learning method called GA\textsuperscript{2}-CLIP. This method effectively improves the generalization from known to unknown categories by introducing hard prompts and generic attribute anchors as a bridge. Our innovations are: The proposed video language fine-tuning generic attribute anchor prompt method can counteract the semantic narrowing problem in the downstream task; we introduce externally supervised hard and soft prompts through a nonlinear mapping layer, which enhances the generalization ability through a competitive learning mechanism. Extensive experiments validate the effectiveness of the method, and we believe that this work provides new research directions in the field of video prompt learning, especially for researchers who lack sufficient experimental conditions.

\bibliographystyle{IEEEtran}

\bibliography{egbib.bib}



\newpage

\vfill

\end{document}